# Fuzzy-Klassen Model for Development Disparities Analysis based on Gross Regional Domestic Product Sector of a Region

Tb. Ai Munandar
Eng. Informatics Dept.
Universitas Serang Raya
Serang – Banten - INDONESIA

Retantyo Wardoyo
Computer Science and Electronic Dept.
Universitas Gajah Mada
Yogyakarta - INDONESIA

## ABSTRACT
Analysis of regional development imbalances quadrant has a very important meaning in order to see the extent of achievement of the development of certain areas as well as the difference. Factors that could be used as a tool to measure the inequality of development is to look at the average growth and development contribution of each sector of Gross Regional Domestic Product (GRDP) based on the analyzed region and the reference region. This study discusses the development of a model to determine the regional development imbalances using fuzzy approach system, and the rules of typology Klassen. The model is then called fuzzy-Klassen. Implications Product Mamdani fuzzy system is used in the model as an inference engine to generate output after defuzzyfication process. Application of MATLAB is used as a tool of analysis in this study. The test a result of Kota Cilegon is shows that there are significant differences between traditional Klassen typology analyses with the results of the model developed. Fuzzy model-Klassen shows GRDP sector inequality Cilegon City is dominated by Quadrant I (K4), where status is the sector forward and grows exponentially. While the traditional Klassen typology, half of GRDP sector is dominated by Quadrant IV (K4) with a sector that is lagging relative status.

## 1. INTRODUCTION
Analysis of development gaps in the study of development economics is basically using Klassen Typology theory by comparing the level of GRDP for the current year with the previous year's, both in terms of total GRDP and GRDP-forming sector itself. Klassen typology theory divides the level of development into four quadrants according to the rules of each based on the comparison of the average value of the growth rate and the average contribution of each sector development, in order to obtain certain quadrants of development (as shown in Table 1 for more details).

This study intends to conduct discussions in order to generate a new approach to analyze the inequality of development of an area using the concept of fuzzy system combined with Klassen typology. The expected outcome of course is able to provide a new approach to analyze the development disparities from the perspective of information technology.

Both methods have advantages in accordance with the position of each concept. Klassen typology widely used by scientists to determine the quadrant of the development of an area by analyzing the existing GRDP sector, for example, to determine the dominant sector economy of a region in order to determine the level of development that has been achieved [1],[2]. So is the concept of fuzzy logic, has been widely implemented in various fields, especially associated with the production planning of a company [3], developing of an auto zoom digital camera [4], the determination of ground water quality [5], assessment of fruit quality for import requirements [6], wormhole detection in computer networks [7], intrusion detection system [8], edge detection for digital image processing [9], forecasting annual rainfall [10], setting temperature and humidity of the room [11] and simulation of the traffic light settings [12]. On this basis then, this research will develop a new model as a tool to analyze the inequality of development of the region by incorporating the concept of fuzzy systems and Klassen typology. This model is expected to carry out the determination of regional development imbalances quadrant based sub-sector gross regional domestic product (GRDP). The model developed later called fuzzy-Klassen.

This study is divided into six sections. The first section discusses the basic background of the research conducted; the second part contains study the literature and theoretical basis used in the study. The third part is a research methodology that includes the steps of research. The fourth section describes the multi-fuzzy system models proposed, the fifth section discusses the results of research and scientific discussions related to testing the model against real case for the determination of the level of development of a region quadrants according to a particular sector. The final section contains conclusions and suggestions for further research in the future.

## 2. LITERATURE REVIEW
### 2.1 Klassen typology
Klassen typology is an analysis tool that is used to describe the growth pattern of economic development of a region based on the comparison of the data of the current year with the previous year or the year specified comparator [1],[2]. Klassen classify the level of development into four main quadrants ie, advanced and rapidly growing sector, depressed growth sector, potential or can still growing sector and relatively underdeveloped sector (see Table 1).

**Table 1. Classification of Economic Growth Matrix According Typology Klassen**

| Quadrant I (K1) | Quadrant II (K2) |
|---|---|
| advanced and rapidly growing sector | depressed growth sector |
| $r_i >= r$ dan $y_i >= y$ | $r_i < r$ dan $y_i >= y$ |
| **Quadrant III (K3)** | **Quadrant IV (K4)** |
| potential or can still growing sector | Relatively underdeveloped sector |
| $r_i >= r$ dan $y_i < y$ | $r_i < r$ dan $y_i < y$ |

Remarks:
$r_i$ : growth rate of GDP sector for district i
$r$ : province GDP growth rate sector
$y_i$ : contribution sector of distrct i
$y$ : province contribution sector

## 2.2 Fuzzy systems

A fuzzy system is a knowledge-based system that contains rule-based or fuzzy IF-THEN rules that contain a statement that is characterized by a membership function (MF) which is continuous. Basically every fuzzy systems will use a different principle from each other while doing a combination of rules, both the pure fuzzy systems, and TSK fuzzy system with fuzyfication and defuzzyfication.

The main problems faced by the pure fuzzy systems is that the input and output of fuzzy sets (in the form of natural language), whereas in engineering science, the input and output of real values of a variable. For this reason, Takagi, Sugeno and Kang (TSK) fuzzy systems proposed in another form, so that the input and the output can be either real values of a variable. While the main problems in the TSK fuzzy systems are, first, THEN part of the form of mathematical equations that do not provide a natural framework for representing knowledge in persons, both, are not given the freedom to use different principles in fuzzy logic, so the usefulness of fuzzy systems that others do not well represented. To overcome the problem of TSK, was proposed three types of fuzzy systems, namely fuzzy system with fuzyfication and defuzzyfication [13].

This study will try to use a third fuzzy model approach. Mamdani inference model of the product is used as a method of analysis to develop a Klassen typology, although it could use models other inference.

## 2.3 Gross Regional Domestic Product

Gross Regional Domestic Product is the value of net output in accordance with the economy of a region of economic activity does (both provincial and district / city) during a specified time period (one calendar year), covering Agriculture, Mining and Quarrying, Manufacturing, Electricity, Gas and Water, Building, Trade, Hotels and Restaurants, Transport and Communication, Finance, Real Estate and Business Services and services.

The calculation of the value of GRDP using two approaches. First approach is direct calculation based on data relevant areas related to the products and services produced. The second approach is using an indirect calculation by adding value-added economic activities and national added value to the group of economic activity area concerned. The results of the calculation, and then presented in the form of calculation of GRDP at current prices and constant prices. In this study, the GRDP at current prices is used as data to be analyzed.

## 3. RESEARCH METHODOLOGY

This study begins with the design of the classification model fuzzy-Klassen. Rule Klassen retained in this model. However, the stages of analysis are used by the ways that are on the concept of fuzzy systems. Here is a diagram of the developed model (as shown in figure 1).

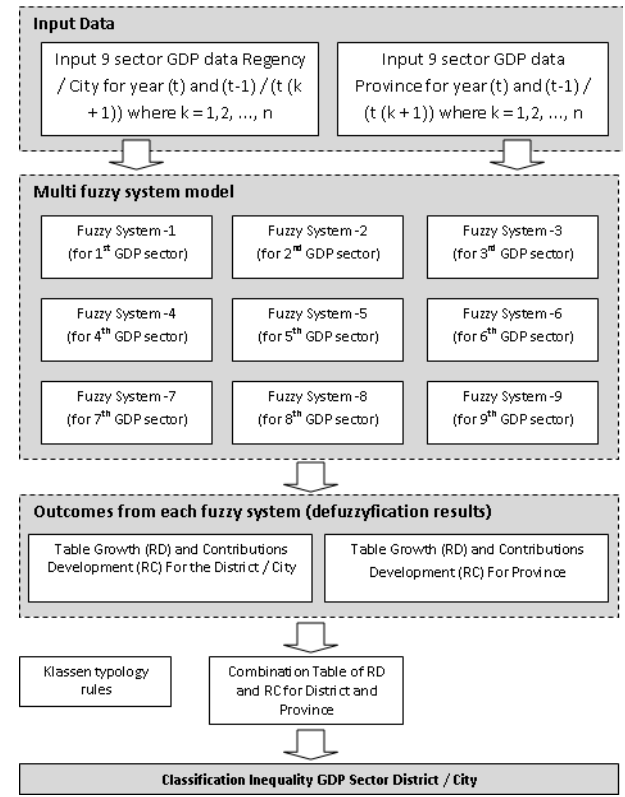

**Figure 1. fuzzy-Klassen model developed**

In the fuzzy model system developed, multi-fuzzy systems are built based on the number of sector GRDP. GRDP sector represents each data input region / district and provincial levels for the current data (t) and the data of previous year (t-1). They are Agriculture ($V_1$), Mining and Quarrying ($V_2$), Manufacturing ($V_3$), Electricity, Gas and Water ($V_4$), Building ($V_5$), Trade, Hotels and Restaurants ($V_6$), Transportation and Communications ($V_7$), Finance, Real Estate and Business Services ($V_8$) and Services ($V_9$). Input as well GDP sector data the district / city and province will be analyzed in the developed fuzzy model. In this model, the fuzzy systems are using parameters such as rules of inference Mamdani Product and defuzzyfication with centroid method. Each fuzzy system will produce an output in the form of an average growth rate of development (RD) and the average contribution of the construction (RC) for the district / city and province. Outcomes fuzzy model system such as RD and RC will be compared between the districts of the province. Klassen typology classification rules are used to form the final classification GDP sector imbalances district / city.

## 4. FUZZY-KLASSEN MODEL DESIGN

Fuzzy-Klassen model in this research is refered to the method of determining GRDP sector imbalances a region with which the Product Mamdani inference used fuzzy rules adopted from Klassen typology. The model developed is also still follow the traditional rules of typology Klassen but data analysis is done with the help of fuzzy systems. Here is a fuzzy model of the system to be used. Fuzzy model-Klassen developed based on the concept of fuzzy rules by adopting rules Klassen typology thus forming composition rules as follows:

$$Ru^l: IF \ V_{k1}(t) \ is \ R_n^l \ AND \ V_{k2}(t-1) \ is \ R_n^l \quad ...(1)$$
$$THEN \ RD \ is P^l \ ALSO \ RC \ is P^l$$

Remarks:
k : indicates the sector to the GDP of k, for k = 1, 2, ..., 9
l : shows the number of rules in the fuzzy basis, for l = 1, 2, ..., M
t : indicates the current GDP sector data
i : variable to - i, for i = 1, 2, ... n
t-1 : shows the GDP of sector data previously or to be compared
V : indicates the sector GDP

For this case, the IF section on fuzzy rules is stated by connecting "AND" (conjunction). While the inference engine used is the Product Mamdani inference engine with the following steps:

**Step 1:** For each fuzzy rules IF - THEN consisting of M rules, count the membership function $\mu_{R_1^l x,...,x\mu_{R_n^l}}(V_{k1},...,V_{ki})$ for (t) and (t 1) for l = 1,2, ..., M

**Step 2:** assume $R_1^l x,..., R_n^l$ as $FP_1$ and $P_1$ as $FP_2$ in the interpretation of fuzzy IF - THEN, then compute
$\mu_{Ru(l)}(V_{ki},...,V_{ki},RD) = \mu_{R_1^l x,...,x\mu_{R_{1n}^l}} \to Rd^l(V_{ki},...,V_{ki},RD)$ And also
$\mu_{Ru(l)}(V_{ki},...,V_{ki},RC) = \mu_{R_1^l x,...,x\mu_{R_{1n}^l}} \to Rd^l(V_{ki},...,V_{ki},RC)$ by using Mamdani implication Product as follows:
$$\mu_{QMM}(V_{ki},...,V_{ki},RD) = \mu_{FP1}(V_{ki},...,V_{ki}), \mu_{FP1}(RD) \ ...(2)$$

For output RD, and,
$$\mu_{QMM}(V_{ki},...,V_{ki},RC) = \mu_{FP1}(V_{ki},...,V_{ki}), \mu_{FP1}(RC) \ ...(3)$$
for output RC.

Step 3: For given input R' in U, compute the fuzzy output Pl' in V for each rule ... using Ponen as follows:
$$\mu_{Bl}'(y) = \sup_{x\in U} t\left[\mu_{A'}(x), \mu_{Ru}(l)(x,y)\right] \quad ...(4)$$

then interpreted into each output as follows: ...(5)

$$\mu_{Pl}'(y) = \sup_{x\in U} t\left[\mu_R(V_{ki}(t)), \mu_R(V_{ki}(t-1)), \mu_{Ru}(l)(V_{ki}(t),V_{ki}(t-1),RD)\right]$$
For RD outcomes and: ...(6)

$$\mu_{Pl}'(y) = \sup_{x\in U} t\left[\mu_R(V_{ki}(t)), \mu_R(V_{ki}(t-1)), \mu_{Ru}(l)(V_{ki}(t),V_{ki}(t-1),RC)\right]$$
for RC outcomes. For l = 1, 2 ... M.

**Step 4:** The output of the fuzzy inference engine is a combination of fuzzy set M {$P_1$, ..., $P_M$} by using a combination of the following:
$$\mu_{B'}(y) = \mu_{B1}'(y) + ,..., +\mu_{BM}'(y) \quad ...(7)$$

then interpreted into each output as follows:
$$\mu_{P'}(RD) = \mu_{B1}'(RD) + ,..., +\mu_{BM}'(RD) \quad ...(8)$$
For RD outcomes, and
$$\mu_{P'}(RC) = \mu_{B1}'(RC) + ,..., +\mu_{BM}'(RC) \quad ...(9)$$
for RC outcomes.

**Step 5:** Defuzzyfication results in step 4 using the centroid method is based on the following equation:

$$y^* = \frac{\int_y y\mu B'(y)dy}{\int_y \mu B'(y)dy} \quad ...(10)$$

Steps 1-5 above applies to any data input GDP indicator consisting of nine variables, for both data and Provincial District. Representations of fuzzy rules are used for each model of fuzzy systems against each indicator of both district and provincial GDP is as follows:

1. $Ru^1: IF \ V_{k1}(t) \ is \ R_1^1 \ AND \ V_{k2}(t-1) \ is \ R_2^1 \ ...(11)$
   $THEN \ RD \ is P^1 \ ALSO \ RC \ is P^1$

2. $Ru^2: IF \ V_{k1}(t) \ is \ R_1^2 \ AND \ V_{k2}(t-1) \ is \ R_2^2 \ ...(12)$
   $THEN \ RD \ is P^2 \ ALSO \ RC \ is P^2$

3. $Ru^3: IF \ V_{k1}(t) \ is \ R_1^3 \ AND \ V_{k2}(t-1) \ is \ R_2^3 \ ...(13)$
   $THEN \ RD \ is P^3 \ ALSO \ RC \ is P^3$

4. $Ru^4: IF \ V_{k1}(t) \ is \ R_1^4 \ AND \ V_{k2}(t-1) \ is \ R_2^4 \ ...(14)$
   $THEN \ RD \ is P^4 \ ALSO \ RC \ is P^4$

5. $Ru^5: IF \ V_{k1}(t) \ is \ R_1^5 \ AND \ V_{k2}(t-1) \ is \ R_2^5 \ ...(15)$
   $THEN \ RD \ is P^5 \ ALSO \ RC \ is P^5$

6. $Ru^6: IF \ V_{k1}(t) \ is \ R_1^6 \ AND \ V_{k2}(t-1) \ is \ R_2^6 \ ...(16)$
   $THEN \ RD \ is P^6 \ ALSO \ RC \ is P^6$

7. $Ru^7: IF \ V_{k1}(t) \ is \ R_1^7 \ AND \ V_{k2}(t-1) \ is \ R_2^7 \ ...(17)$
   $THEN \ RD \ is P^7 \ ALSO \ RC \ is P^7$

8. $Ru^8: IF \ V_{k1}(t) \ is \ R_1^8 \ AND \ V_{k2}(t-1) \ is \ R_2^8 \ ...(18)$
   $THEN \ RD \ is P^8 \ ALSO \ RC \ is P^8$

9. $Ru^9: IF \ V_{k1}(t) \ is \ R_1^9 \ AND \ V_{k2}(t-1) \ is \ R_2^9 \ (19)$
   $THEN \ RD \ is P^9 \ ALSO \ RC \ is P^9$

The fuzzy rules applicable to each indicator GDP on fuzzy model system developed.

## 5. RESULTS AND DISCUSION
Tests carried out on the data model of Kota Cilegon GRDP sector for 2012 and 2013 (as shown in table 2).

**Table 2. Input data for each fuzzy system**

| Sector | 2011(t-1) | 2012(t) |
|---|---|---|
| $V_1$ | 293.563,49 | 296.121,45 |
| $V_2$ | 12.101,38 | 12.935,68 |
| $V_3$ | 13.218.285,53 | 14.107.542,93 |
| $V_4$ | 980.774,99 | 1.010.756,92 |
| $V_5$ | 60.863,74 | 65.161,53 |
| $V_6$ | 2.139.891,00 | 2 357.486,68 |
| $V_7$ | 907.932,85 | 951.873,51 |
| $V_8$ | 408.769,00 | 442.926,24 |
| $V_9$ | 206.107,98 | 225.763,38 |

In this paper, discussed only one fuzzy system processes the V1 sector to illustrate the developed model. The next process to the data V1 to V9 is conducted in the same way. The first step, the data for the indicator V1 then inserted into the membership function V1 as follows:

$$\mu_{low} = \begin{cases} 1 & x \leq 216.831 \\ (283.777-x)/66.945 & 216.831 \leq x \leq 283.777 \\ 0 & x \geq 283.777 \end{cases} \quad ...(20)$$

$$\mu_{medium} = \begin{cases} 1 & x = 283{,}777 \\ (x - 216.831)/66.945 & 216{,}831 \leq x \leq 283{,}777 \\ (350{,}722 - x)/66.945 & 283{,}777 \leq x \leq 350{,}722 \\ 0 & x \leq 216{,}832; x \geq 350{,}722 \end{cases} \quad \ldots (21)$$

$$\mu_{high} = \begin{cases} 1 & x \geq 350{,}722 \\ (x - 283{,}777)/66.945 & 283{,}777 \leq x \leq 350{,}722 \\ 0 & x \leq 283{,}777 \end{cases} \quad \ldots (22)$$

Equation (20), (21) and (22) applies to years of data (t) and (t-1) for the district and province. Based on the data in Table 2 above, the value of V1 for (t) is equal to 296,121.45 and (t-1) of 293,563.49. Fuzyfication results for $V_1$ (t) and $V_1$ (t-1) are as follows:

- Agricultural Sector V1 (t)
  $\mu_{low} = 0$; $\mu_{medium} = 0.816$; $\mu_{high} = 0.184$

- Agricultural Sector V1 (t-1)
  $\mu_{low} = 0$; $\mu_{medium} = 0.854$; $\mu_{high} = 0.146$

The next step, for each outcome variable RD and RC, do fuzzyfication for each point. Here are the results fuzzyfication any outcome from the point of 1-100:

- Average growth (RD)

**μLow :**
For $x \leq 30$, then $\mu_{low}$ (RD) = 1
For $30 \leq x \leq 80$ then $\mu_{low}$ (RD) obtained based on following point:
$\mu_{low}$ (31) = 0.9796 ; $\mu_{low}$ (32) = 0.9592 ; $\mu_{low}$ (33) = 0.9388;
…. Etc.
$\mu_{low}$ (80) = 0;
For $x \geq 80$, then $\mu_{low}$ (RD) = 0;

**μHigh :**
For $x \leq 30$, then $\mu_{high}$ (RD) = 0;
For $30 \leq x \leq 80$ then $\mu_{high}$ (RD) obtained based on following point:
$\mu_{high}$ (31) = 0.0204; $\mu_{high}$ (32) = 0.0408; $\mu_{high}$ (33) = 0.0612;
…. Etc
$\mu_{high}$ (80) = 0;
For $x \geq 80$, then $\mu_{high}$ (RD) = 1;

- Contribution rate of the development (RC)

**μLow:**
for $x \leq 33$, then $\mu_{low}$ (RC) = 1;
for $33 \leq x \leq 73$ then $\mu_{low}$ (RC) obtained based on following point:
$\mu_{low}$ (34) = 0.9744; $\mu_{low}$ (35) = 0.9487; $\mu_{low}$ (36) = 0.9231;
…. Etc
$\mu_{low}$ (73) = 0;
For $x \geq 73$, then $\mu_{low}$ (RC) = 0;

**μHigh:**
For $x \leq 33$, then $\mu_{high}$ (RC) = 0;
For $33 \leq x \leq 73$ then obtained based on following point:
$\mu_{high}$ (34) = 0.0029; $\mu_{high}$ (35) = 0.0059; $\mu_{high}$ (36) = 0.0088;
…. Etc.
$\mu_{high}$ (73) = 0;
For $x \geq 73$, then $\mu_{high}$ (RC) = 1;

By entering fuzzy rules from equation (11) to (19) into equation (5) for the outcomes RD and (6) for the output RC, further processing rules based on the data that has been fuzzyficated above. Here are some discussion fuzzyfication processing results into the rules for agriculture ($V_1$):

R1 (rule 1) : IF V1(t) = Low (R) AND V1(t-1) = Low (R) THEN RD($V_i$) = Low (R) ALSO RC($V_i$)= Low (R)

- **For RD outcomes**

**Point 1 - 30**

$$\mu_{FR1}(V_1(t), V_1(t-1), RD) = \min\left[\min\begin{bmatrix} \mu_{low}(V_1(t)), \\ \mu_{low}(V_1(t-1)) \end{bmatrix}, \mu_{low}(RD)\right]$$

$$\mu_{FR1}(V_1(t), V_1(t-1), RD) = \min\left[\min\begin{bmatrix} \mu_{low}(296.12), \\ \mu_{low}(293.56) \end{bmatrix}, \mu_{low}(1)\right]$$

$$\mu_{FR1}(V_1(t), V_1(t-1), RD) = \min\left[\min\begin{bmatrix} \mu_{low}(0), \\ \mu_{low}(0) \end{bmatrix}, \mu_{low}(1)\right] = 0$$

$$\mu_{FR1}(V_1(t), V_1(t-1), RD) = \min\left[\min\begin{bmatrix} \mu_{low}(296.12), \\ \mu_{low}(293.56) \end{bmatrix}, \mu_{low}(2)\right]$$

$$\mu_{FR1}(V_1(t), V_1(t-1), RD) = \min\left[\min\begin{bmatrix} \mu_{low}(0), \\ \mu_{low}(0) \end{bmatrix}, \mu_{low}(1)\right] = 0$$

….. Etc.

**Point 31 - 80**

$$\mu_{FR1}(V_1(t), V_1(t-1), RD) = \min\left[\min\begin{bmatrix} \mu_{low}(V_1(t)), \\ \mu_{low}(V_1(t-1)) \end{bmatrix}, \mu_{low}(RD)\right]$$

$$\mu_{FR1}(V_1(t), V_1(t-1), RD) = \min\left[\min\begin{bmatrix} \mu_{low}(296.12), \\ \mu_{low}(293.56) \end{bmatrix}, \mu_{low}(31)\right]$$

$$\mu_{FR1}(V_1(t), V_1(t-1), RD) = \min\left[\min\begin{bmatrix} \mu_{low}(0), \\ \mu_{low}(0) \end{bmatrix}, \mu_{low}(0.9796)\right] = 0$$

$$\mu_{FR1}(V_1(t), V_1(t-1), RD) = \min\left[\min\begin{bmatrix} \mu_{low}(296.12), \\ \mu_{low}(293.56) \end{bmatrix}, \mu_{low}(32)\right]$$

$$\mu_{FR1}(V_1(t), V_1(t-1), RD) = \min\left[\min\begin{bmatrix} \mu_{low}(0), \\ \mu_{low}(0) \end{bmatrix}, \mu_{low}(0.9592)\right] = 0$$

….. Etc.

**Point 81 - 100**

$$\mu_{FR1}(V_1(t), V_1(t-1), RD) = \min\left[\min\begin{bmatrix} \mu_{low}(V_1(t)), \\ \mu_{low}(V_1(t-1)) \end{bmatrix}, \mu_{low}(RD)\right]$$

$$\mu_{FR1}(V_1(t), V_1(t-1), RD) = \min\left[\min\begin{bmatrix} \mu_{low}(296.12), \\ \mu_{low}(293.56) \end{bmatrix}, \mu_{low}(81)\right]$$

$$\mu_{FR1}(V_1(t), V_1(t-1), RD) = \min\left[\min\begin{bmatrix} \mu_{low}(0), \\ \mu_{low}(0) \end{bmatrix}, \mu_{low}(0)\right] = 0$$

$$\mu_{FR1}(V_1(t), V_1(t-1), RD) = \min\left[\min\begin{bmatrix} \mu_{low}(296.12), \\ \mu_{low}(293.56) \end{bmatrix}, \mu_{low}(82)\right]$$

$$\mu_{FR1}(V_1(t), V_1(t-1), RD) = \min\left[\min\begin{bmatrix} \mu_{low}(0), \\ \mu_{low}(0) \end{bmatrix}, \mu_{low}(0)\right] = 0$$

….. Etc.

- **For RC outcomes**

**Point 1 - 30**

$$\mu_{FR1}(V_1(t), V_1(t-1), RC) = \min\left[\min\begin{bmatrix} \mu_{low}(V_1(t)), \\ \mu_{low}(V_1(t-1)) \end{bmatrix}, \mu_{low}(RC)\right]$$

$$\mu_{FR1}(V_1(t), V_1(t-1), RC) = \min\left[\min\begin{bmatrix} \mu_{low}(296.12), \\ \mu_{low}(293.56) \end{bmatrix}, \mu_{low}(1)\right]$$

$$\mu_{FR1}(V_1(t), V_1(t-1), RC) = \min\left[\min\begin{bmatrix} \mu_{low}(0), \\ \mu_{low}(0) \end{bmatrix}, \mu_{low}(1)\right] = 0$$

$$\mu_{FR1}(V_1(t), V_1(t-1), RC) = \min\left[\min\begin{bmatrix}\mu_{low}(296.12),\\ \mu_{low}(293.56)\end{bmatrix}, \mu_{low}(2)\right]$$

$$\mu_{FR1}(V_1(t), V_1(t-1), RC) = \min\left[\min\begin{bmatrix}\mu_{low}(0),\\ \mu_{low}(0)\end{bmatrix}, \mu_{low}(1)\right] = 0$$

….. Etc.

**Point 31 - 80**

$$\mu_{FR1}(V_1(t), V_1(t-1), RC) = \min\left[\min\begin{bmatrix}\mu_{low}(V_1(t)),\\ \mu_{low}(V_1(t-1))\end{bmatrix}, \mu_{low}(RC)\right]$$

$$\mu_{FR1}(V_1(t), V_1(t-1), RC) = \min\left[\min\begin{bmatrix}\mu_{low}(296.12),\\ \mu_{low}(293.56)\end{bmatrix}, \mu_{low}(31)\right]$$

$$\mu_{FR1}(V_1(t), V_1(t-1), RC) = \min\left[\min\begin{bmatrix}\mu_{low}(0),\\ \mu_{low}(0)\end{bmatrix}, \mu_{low}(0.9744)\right] = 0$$

$$\mu_{FR1}(V_1(t), V_1(t-1), RC) = \min\left[\min\begin{bmatrix}\mu_{low}(296.12),\\ \mu_{low}(293.56)\end{bmatrix}, \mu_{low}(32)\right]$$

$$\mu_{FR1}(V_1(t), V_1(t-1), RC) = \min\left[\min\begin{bmatrix}\mu_{low}(0),\\ \mu_{low}(0)\end{bmatrix}, \mu_{low}(0.9487)\right] = 0$$

….. Etc.

**Point 81 - 100**

$$\mu_{FR1}(V_1(t), V_1(t-1), RC) = \min\left[\min\begin{bmatrix}\mu_{low}(V_1(t)),\\ \mu_{low}(V_1(t-1))\end{bmatrix}, \mu_{low}(RC)\right]$$

$$\mu_{FR1}(V_1(t), V_1(t-1), RC) = \min\left[\min\begin{bmatrix}\mu_{low}(296.12),\\ \mu_{low}(293.56)\end{bmatrix}, \mu_{low}(81)\right]$$

$$\mu_{FR1}(V_1(t), V_1(t-1), RC) = \min\left[\min\begin{bmatrix}\mu_{low}(0),\\ \mu_{low}(0)\end{bmatrix}, \mu_{low}(0)\right] = 0$$

$$\mu_{FR1}(V_1(t), V_1(t-1), RC) = \min\left[\min\begin{bmatrix}\mu_{low}(296.12),\\ \mu_{low}(293.56)\end{bmatrix}, \mu_{low}(82)\right]$$

$$\mu_{FR1}(V_1(t), V_1(t-1), RC) = \min\left[\min\begin{bmatrix}\mu_{low}(0),\\ \mu_{low}(0)\end{bmatrix}, \mu_{low}(0)\right] = 0$$

….. Etc.

For all the rest of the fuzzy rules also apply the same thing as the above rule processing step. This also applies to the eight other indicators in the fuzzy model system developed. Based on the third proces above, we obtaining Rule 9 (R9) agree for inference and defuzyfication steps.

The next step then is performing the calculation outcomes of RD and RC by equation (8) and (9) and then determine the final value defuzyfication by equation (10). We obtained ooutcomes RD and RC as follow :

**For RD outcomes,**
$\mu_{P'}(RD) = \max[\mu_{P1'}(RD), ... , \mu_{PM'}(RD)]$
$\mu_{P'}(RD) = \max[\mu_{P1'}(0), ... , \mu_{P50'}(0.515), ... , \mu_{P100'}(1)]$;
$\mu_{P'}(RD) = 1$;

**For RC outcomes,**
$\mu_{P'}(RC) = \max[\mu_{P1'}(RC), ... , \mu_{PM'}(RC)]$
$\mu_{P'}(RC) = \max[\mu_{P1'}(0), ... , \mu_{P50'}(0.388), ... , \mu_{P100'}(1)]$;
$\mu_{P'}(RC) = 1$;

The final step is to defuzycifation for each point of output (RD and RC). Centroid method as in Equation 10 is used at this stage. Results defuzycifation each output RD and RC was obtained as follows:

$$RD = \frac{1x1+,\cdots,+1x25+,\cdots,+1x50+,\cdots,+1x75+,\cdots,+1x100}{1+,\cdots,+1+,\cdots,+1+,\cdots,+1}$$

RD = 37,79;

$$RD = \frac{1x1+,\cdots,+1x35+,\cdots,+1x60+,\cdots,+1x85+,\cdots,+1x100}{1+,\cdots,+1+,\cdots,+1+,\cdots,+1}$$

RC = 35,38

Defuzzyfication procesess for V1 indicator Results obtained of 37.79 for RD and of 35.38 for RC. Table 3 shows the results of steps 1-5 for the other eight sectors GDP.

**Table 3. Defuzzification results of GDP for Kota Cilegon**

| Sector | RD | RC |
|---|---|---|
| $V_1$ | 37.8 | 35.4 |
| $V_2$ | 75.5 | 75.3 |
| $V_3$ | 75.5 | 75.3 |
| $V_4$ | 32.1 | 30.8 |
| $V_5$ | 29.1 | 27.5 |
| $V_6$ | 75.5 | 75.3 |
| $V_7$ | 30.9 | 29.5 |
| $V_8$ | 75.5 | 75.3 |
| $V_9$ | 75.5 | 75.3 |

In the same way the earlier stages of fuzzy systems, RD and RC values calculated for the province, in order to obtain the following results (as shown in Table 4).

**Table 4. Defuzzification results of GRDP Banten Province**

| Sector | RD | RC |
|---|---|---|
| $V_1$ | 29.1 | 27.5 |
| $V_2$ | 29.1 | 27.5 |
| $V_3$ | 39.8 | 39.2 |
| $V_4$ | 29.1 | 27.5 |
| $V_5$ | 29.1 | 27.5 |
| $V_6$ | 75.5 | 75.3 |
| $V_7$ | 75.5 | 75.3 |
| $V_8$ | 29.1 | 27.5 |
| $V_9$ | 75.5 | 75.3 |

The next step is to compare the value of RD and RC Province against the District. In this step, Klassen rules as shown in Table 1 are used to compare the value of RD and RC between districts / cities in the province. Classification according to the development gaps GDP indicator as follows (see Table 5):

**Table 5. Disparities classification of development by GRDP sector of Kota Cilegon using Fuzzy-Klassen**

| Sector | Kota Cilegon | | Banten Province | | Quadrant |
|---|---|---|---|---|---|
| $V_1$ | 29.1 | 27.5 | 29.1 | 27.5 | K1 |
| $V_2$ | 75.5 | 75.3 | 29.1 | 27.5 | K1 |
| $V_3$ | 75.5 | 75.3 | 39.8 | 39.2 | K1 |
| $V_4$ | 32.1 | 30.8 | 29.1 | 27.5 | K1 |
| $V_5$ | 29.1 | 27.5 | 29.1 | 27.5 | K1 |
| $V_6$ | 75.5 | 75.3 | 75.5 | 75.3 | K1 |
| $V_7$ | 30.9 | 29.5 | 75.5 | 75.3 | K4 |
| $V_8$ | 75.5 | 75.3 | 29.1 | 27.5 | K1 |
| $V_9$ | 75.5 | 75.3 | 75.5 | 75.3 | K1 |

Based on Table 5 it can be seen that the Agriculture sector ($V_1$) of Kota Cilegon is in Quadrant I (K1) which sectors are developed and grown by leaps and bounds. Likewise with the other sectors of the Mining and Quarrying ($V_2$), Manufacturing ($V_3$), Electricity, Gas and Water (V4), Building ($V_5$), Trade, Hotels and Restaurants ($V_6$), Finance, Real Estate and Business Services ($V_8$) and services ($V_9$). Only the transport and communications sector ($V_7$) which should be the focus in the future. The results of the analysis with Fuzzy-Klassen shows that the transport and communications sector ($V_7$) enter into Quadrant IV (K4) which is relatively underdeveloped sector.

## 6. CONCLUSION

This research was modeling the development disparities using sector GDP by combining multi-fuzzy system models and rules Klassen typology. The test results in Kota Cilegon showed that in general, the model is able to analyze the inequality sector GRDP of a region. However, this model still requires further studies related to the fuzzy rule used in the process of step 2 and 3 of the model developed. Results obtained between fuzzy-Klassen with traditional Klassen typology showed a highly significant difference. In the model developed, dominated by the inequality Quadrant I (K1) while the traditional typology Klassen shows Quadrant diverse, but dominated by Quadrant IV (K4). The cause of this difference may occur due to the adoption of rules of Klassen less precise, so that the resulting fuzzy rule is still not quite right.

Subsequent studies, models can be developed by combining decision tree to extract rules from the data classification based typology Klassen using national data. The use of decision tree for forming the rules are expected to form a fuzzy rule is better than the rule in use today. Reinforcement-strengthening the model developed is still needed to enhance the fuzzy-Klassen later.

## 7. REFERENCES

[1] Fachrurrazy, 2009, Analisis Penentuan Sektor Unggulan Perekonomian Wilayah Kabupaten Aceh Utara Dengan Pendekatan Sektor Pembentuk PDRB. *Tesis*, Sekolah Pascasarjana - Universitas Sumatera Utara. *In bahasa.*

[2] Sudarti, 2009. Penentuan Leading Sektor Pembangunan Daerah Kabupaten/Kota Di Jawa Timur, *Jurnal HUMANITY*, Volume V, Nomor 1, September 2009: 68 – 79. *In bahasa.*

[3] Firmansyah, Iman dan Silvia Firda Utami, 2013. Tsukamoto Fuzzy Logic Application in Production Planning at PT. Kimia Farma (Persero) Tbk. Plant Bandung Indonesia, *Proceedings The 2nd International Conference On Global Optimization and Its Applications 2013 (ICoGOIA2013)*

[4] Vasant, I. Elamvazuthi, P. and J.Webb, 2009. The Application of Mamdani Fuzzy Model for Auto Zoom Function of a Digital Camera, (*IJCSIS) International Journal of Computer Science and Information Security*, Vol. 6, No. 3, 2009, pp. 244 – 249

[5] Nasr, A. Saberi., M. Rezaei and M. Dashti Barmaki, 2012. Analysis of Groundwater Quality using Mamdani Fuzzy Inference System (MFIS) in Yazd Province, Iran, *International Journal of Computer Applications* (0975 – 8887), Volume 59– No.7, December 2012, pp. 45 – 53

[6] Alavi, N, 2012. Date Grading Using Rule-Based Fuzzy Inference System, *Journal of Agricultural Technology* 2012 Vol. 8(4), pp. 1243-1254

[7] Revathi,P., M. M. Sahana and Vydeki Dharmar, 2013. Cross Layer Detection of Wormhole In MANET Using FIS, *Journal of ITSI Transactions on Electrical and Electronics Engineering (ITSI-TEEE),* Volume -1, Issue - 3, 2013, pp. 75 – 79

[8] Harikishan, ANVBS., and P.Srinivasulu, 2013. Intrusion Detection System Using Fuzzy Inference System, *International Journal of Computer & Organization Trends* –Volume 3 Issue 8 – Sep 2013, pp. 345 - 352

[9] Yadav, Meenakshi and Kalpna Kashyap, 2013. Edge Detection Through Fuzzy Inference System, *International Journal Of Engineering And Computer Science,* Volume 2 Issue 6 June, 2013, pp. 1855-1860

[10] Fallah-Ghalhary, G.A., Mousavi-Baygi, M and Nokhandan, M.H, 2009. Annual Rainfall Forecasting by Using Mamdani Fuzy Inference System, *Research Journal of Environmental Sciences* 3 (4) : 400 - 413

[11] Das, Tarun Kumar and Das, Y, 2013. Design of A Room Temperature And Humidity Controller Using Fuzzy Logic, *American Journal of Engineering Research (AJER)* , Volume-02, Issue-11, pp-86-97

[12] Tan, Kok Khiang., Khalid, N and Yusof, R, 1996. Intelligent Traffic Lights Control By Fuzzy Logic, *Malaysian Journal of Computer Science*, Vol. 9 No. 2, December 1996, pp. 29-35.

[13] Wang, Li-Xin, 1995, *A Course in Fuzzy Systems and Control*, Prentice-Hall International, Inc